\documentclass[runningheads]{llncs}
\usepackage{graphicx}
\usepackage{algorithmic}
\usepackage{multirow}
\begin{document}
\title{Comparison of Automated Machine Learning Tools for SMS Spam Message Filtering}
\author{Waddah Saeed %\orcidID{0000-0002-2280-4427}
}
\authorrunning{W. Saeed}

\institute{Center for Artificial Intelligence Research (CAIR), University of Agder, Jon Lilletuns vei 9, 4879 Grimstad, Norway \\
\email{waddah.waheeb@uia.no}}
\maketitle              % typeset the header of the contribution
\begin{abstract}
Short Message Service (SMS) is a very popular service used for communication by mobile users. However, this popular service can be abused by executing illegal activities and influencing security risks. Nowadays, many automatic machine learning (AutoML) tools exist which can help domain experts and lay users to build high-quality ML models with little or no machine learning knowledge. In this work, a classification performance comparison was conducted between three automatic ML tools for SMS spam message filtering. These tools are mljar-supervised AutoML, H2O AutoML, and Tree-based Pipeline Optimization Tool (TPOT) AutoML. Experimental results showed that ensemble models achieved the best classification performance. The Stacked Ensemble model, which was built using H2O AutoML, achieved the best performance in terms of Log Loss (0.8370), true positive (1088/1116), and true negative (281/287) metrics. There is a 19.05\% improvement in Log Loss with respect to TPOT AutoML and 5.56\% improvement with respect to mljar-supervised AutoML. The satisfactory filtering performance achieved with AutoML tools provides a potential application for AutoML tools to automatically determine the best ML model that can perform best for SMS spam message filtering.

\end{abstract}
\section{Introduction}

Short Message Service (SMS) is a very popular service that enables its users to send short text messages from one mobile device to another. However, mobile users can receive SMS spam messages. 

According to \cite{ABAYOMIALLI2019197,waheeb2017content}, the huge number of mobile devices/users and the possibility of sending bulk SMS messages easily and with low cost are factors that contribute to the growth of SMS spam problem and attract malicious organizations for executing illegal activities and influencing security risks.

Content-based filtering has been extensively studied to combat SMS spam messages. This type of filtering uses techniques to analyze selected features extracted from SMS messages with the aim to filter spam messages.

Various machine learning models have been utilized for SMS spam message filtering such as  support vector machine (SVM) \cite{almeida2013towards,almeida2011contributions}, multilayer perceptron \cite{waheeb2015content}, and deep learning \cite{fi12090156,ROY2020524}.

Nowadays, many automatic ML tools (AutoML) exist. With these AutoML tools, domain experts are enabled to build ML applications without extensive knowledge of statistics and machine learning \cite{zoller2021benchmark}. Furthermore, lay users with little or no ML knowledge can use user-friendly automated systems to build high-quality custom models \cite{he2021automl}. In the literature of SMS spam message filtering, one work used an AutoML tool to make a classification performance comparison between various ML models built using that AutoML tool \cite{SULEIMAN2017154}. 

Clearly, to the best of our knowledge, there is no study comparing SMS spam message filtering performance of the best models built using AutoML tools. Therefore, this work carried out a comparison between three AutoML tools for SMS spam message filtering. These tools are mljar-supervised AutoML \cite{mljar}, H2O AutoML \cite{H2OAutoML20}, and Tree-based Pipeline Optimization Tool (TPOT) AutoML \cite{le2020scaling}. The importance of this work is to investigate how good the classification performance using the three selected tools, how fast is the training process, and how much difference in the performance between these three tools.

The remainder of this paper is organized as follows. Related work is given in Section 2. Experimental settings are described in Section 3. Results and discussions are given in Section 4. Finally, conclusions and possible future works are highlighted in the last section.

\section{Related Works}

Content-based SMS spam filtering uses techniques to analyse selected features extracted from SMS messages with the aim to filter spam messages.

In \cite{almeida2013towards,almeida2011contributions}, various classifiers were used to classify SMS messages. These classifiers were naive Bayes, C4.5, k-nearest neighbors, and SVM. Two tokenizers were used by the authors. In the first tokenizer, alphanumeric characters followed a printable character.Dots, commas, and colons were excluded from the middle of the pattern. The second tokenizer was represented by any sequence of characters separated by dots, commas, colons, blanks, tabs, returns, and dashes. It was found that better performance was achieved using the first tokenizer with accuracy equals 97.64\%.

In \cite{goswami2016automated}, stylistic and text features were utilized with two SVM classifiers to filter SMS spam messages. An SMS message was classified as a spam message if both SVM classifiers classified the message as a spam message. It was found that the proposed methodology with two SVM classifiers was better than using one SVM classifier.

Multilayer perceptron with features selected by the Gini index (GI) method was used in \cite{waheeb2015content}. According to the obtained results, the best AUC performance was around 0.9648, which was achieved with 100-features.

In \cite{waheeb2017content}, a classification performance comparison was conducted between ten feature subset sizes which were selected by three feature selection methods. SVM was used as a classifier and trained with the feature subset sizes selected by feature selection methods. Based on the obtained results, the features selected by information gain (IG) enhanced the classification performance of the SVM classifier with the ten feature subset sizes. The best result was achieved with only 50\% of the extracted features. Based on that, it was concluded that the feature selection step should be used because using a big number of features as inputs could lead to degrading the classification performance.

In \cite{app10145011} the authors proposed a method based on the discrete hidden Markov model for two reasons. The first one is to use the word order information and the second reason to solve the low term frequency issue found in SMS messages. This proposed method scored 0.959 in terms of accuracy.

Deep learning models were used for SMS spam message filtering, for example, the works in \cite{fi12090156,ROY2020524}. In \cite{fi12090156}, the authors proposed a hybrid deep learning model based on the combination of Convolutional Neural Network (CNN) and Long Short-Term Memory (LSTM). The classifier was developed to deal with SMS messages that are written in Arabic or English. It was found that the proposed CNN-LSTM model outperformed several machine learning classifiers with an accuracy of 98.37\%. The authors in \cite{ROY2020524} used CNN and LSTM models for classification. It was found that both models achieved higher classification accuracy compared to other ML models, with 3 CNN + Dropout being the most accurate model achieving an accuracy of 99.44\%.

Nowadays, many automatic ML tools exist such as mljar-supervised AutoML \cite{mljar}, H2O AutoML \cite{H2OAutoML20}, and Tree-based Pipeline Optimization Tool (TPOT) AutoML \cite{le2020scaling}. Domain experts can benefit from such AutoML tools because AutoML tools can enable them to build ML applications without extensive knowledge of statistics and machine learning \cite{zoller2021benchmark}. Furthermore, lay users with little or no ML knowledge can use user-friendly automated systems to build high-quality custom models \cite{he2021automl}. According to \cite{he2021automl}, there are four main steps in the AutoML pipeline: data preparation, feature engineering, model generation, and model evaluation. In data preparation step, the given data is prepared to be used to train and test ML models. In the second step, a dynamic combination of feature extraction, feature construction, and feature selection processes are used to come up with useful features that can be used by ML models. Search space and optimization methods are two main components in model generation step. The last step in the AutoML pipeline is evaluating the built ML models.

In the literature of SMS spam message filtering, one work used H2O AutoML to make a classification performance comparison between various ML models \cite{SULEIMAN2017154}. Based on the obtained results, it was found that the number of digits and existing of URL in SMS messages are the most significant features that contribute highly to detect SMS spam messages. It was also found that random forest is the best model for the used dataset with 0.977\% in terms of accuracy.

Clearly, to the best of our knowledge, there is no study comparing SMS spam message filtering performance of the best models built using AutoML tools. Therefore, this work investigated the abilities of three AutoML tools for SMS spam message filtering.

\section{Methodology}
The methodology consists of two main steps: data collection and the setting used with the three automatic machine learning tools.

\subsection{Data Used}
In this work, the data used in the simulations is the post-processed data used in ~\cite{waheeb2017content}\footnote{https://github.com/Waddah-Saeed/EnglishSMSCollection/blob/master/IG.zip}. In ~\cite{waheeb2017content}, the data was collected from three works \cite{almeida2011contributions,delany2012sms,nuruzzaman2011independent} then pre-processed by removing duplicate messages and non-English messages. The number of messages after the removal is 5,610 messages: 4,375 legitimate messages and 1,235 spam messages. Following that, as explained in ~\cite{waheeb2017content}, text pre-processing methods were used to reduce the number of extracted features including lowercase conversion, feature abstraction replacement, tokenization, and stemming.

There are 6,463 features in the data set. Therefore, in this work, features selected using the information gain (IG) method were used. IG was selected because its selected features helped SVM to achieve better results as reported in ~\cite{waheeb2017content}.

In this work, 25\% of the data set was used as a test set. The data set was split in a stratified fashion. Three feature subset sizes were selected with sizes equal to 50, 100, and 200.

\subsection{Settings Used with the Automatic Machine Learning Tools}
The settings used with the automatic machine learning tools used in this work are described below.

\subsubsection{mljar-supervised AutoML} It is an automated ML tool that works with tabular data \cite{mljar}. Various ML models can be selected to be used for classification or regression tasks. In this work, nine models were used namely Baseline, Decision Tree, Random Forest, Xgboost, LightGBM, CatBoost, Extra Trees, Neural Network, and Nearest Neighbors.

The mljar-supervised AutoML has several steps that can be used in the process of searching for the best performing ML model in the ML pipeline. Not all ML models can be used in these steps. In this work, the steps with the ML model used are given below:

\begin{enumerate}
    \item Using Baseline and Decision Tree models to get quick insights from the data.
    \item The selected models except for Baseline and Decision Tree were trained with default hyperparameters.
    \item Random Search step was used over a defined set of hyperparameters with the seven models in Step 2.
    \item Golden Features (i.e., new-constructed features) were used with Xgboost, LightGBM, and CatBoost models.
    \item  New models based on Random Forest, Xgboost, LightGBM, CatBoost, Extra Trees, and Neural Network were trained on selected features.
    \item The top two performing models were tuned further in what is called a hill-climbing step.
    \item The last step is the ensemble step where all models from the previous steps were ensembled.
\end{enumerate}

In this work, the command used to initialize AutoML object in mljar-supervised AutoML (version 0.10.6) is shown in Fig. \ref{fig1}:

\begin{figure}
\includegraphics[width=\textwidth]{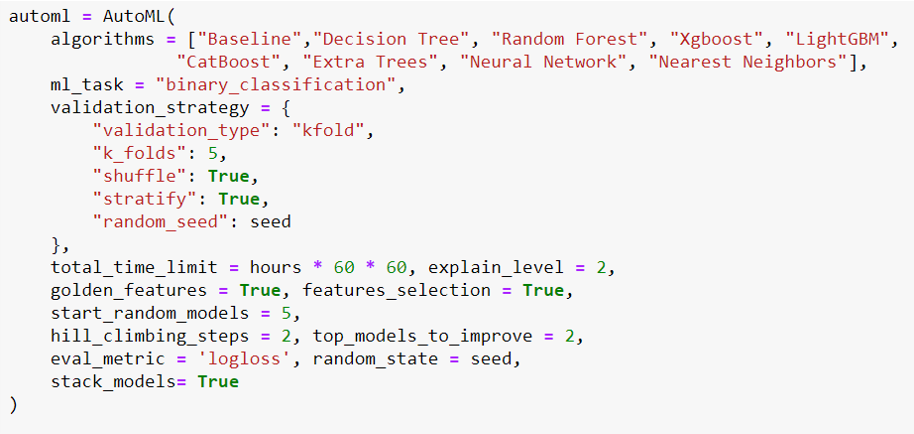}
\caption{Initialize AutoML object in mljar-supervised AutoML.} \label{fig1}
\end{figure}

\subsubsection{H2O’s AutoML} 
\begin{figure}
\includegraphics[width=\textwidth]{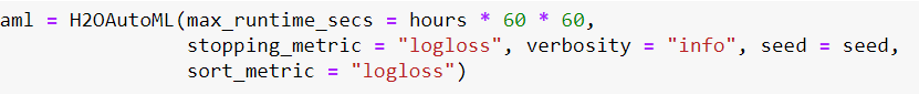}
\caption{Initialize AutoML object in H2O AutoML.} \label{fig2}
\end{figure}

With H2O AutoML \cite{H2OAutoML20}, various ML models based on Random Forest, Generalized Linear Model (GLM), Gradient Boosting Machine (GBM), Deep Learning (a fully-connected multi-layer neural network) can be built. In this work, the execution steps started with using the default settings with XGBoost, GLM, Random Forest, and Deep Learning models. Then, built an Extremely Randomized Trees (XRT) model. Following that, a grid search was used with XGBoost, GBM, and Deep Learning models. After that, learning rate annealing with GBM and learning rate search with XGBoost were used. Finally, two Stacked Ensembles were built. The code used to initialize AutoML object in H2O AutoML (version 3.32.1.3) is shown in Fig. \ref{fig2}.

\subsubsection{TPOT} 
\begin{figure}
\includegraphics[width=\textwidth]{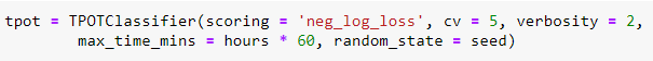}
\caption{Initialize AutoML object in TPOT.} \label{fig3}
\end{figure}

TPOT AutoML \cite{le2020scaling} is a tool that optimizes ML pipelines using genetic programming. Various ML models and their variations are evaluated by TPOT namely Naive Bayes, Decision Tree, Extra Trees, Random Forest, Gradient Boosting, Logistic Regression, Xgboost, Neural Network, and Nearest Neighbors. The code used to initialize AutoML object in TPOT (version 0.11.7) is shown in Fig. \ref{fig3}.

\subsubsection{General Settings} For a fair comparison between the tools, the following settings were used with the three tools:

\begin{itemize}
    \item Cross-validation settings: Five folds were created with a stratified fashion. However, how the samples in these folds were selected from the training data are controlled by the tools.
    \item Performance metric: Log Loss was used as a performance metric.
    \item Training time: Two, three, and four hours were given to train the models using 50, 100, and 200 features, respectively.
\end{itemize}

\section{Results and Discussion}

This section reports and discusses the obtained results using the given methodology in the previous section. It starts with the classification performance comparison using three different feature subset sizes. Following that, the best classification performance for each feature size and for each tool are presented and discussed.

\subsection{Classification performance comparison}
The best model on the training set from each tool was used for prediction. The performance of the best model in each tool is shown in Table \ref{tab1}. It can be seen from Table \ref{tab1} that the best model is the Gradient Boosting model built using the TPOT AutoML tool. The worst performance was using a Deep Learning model built using the H2O AutoML tool. The bad Log Loss value obtained by the Deep Learning model is because of the incorrect prediction as shown in the false positive component in Table \ref{tab1}. 

Turning now to the results in Table \ref{tab2} and Table \ref{tab3}, it seems that increasing the number of features helped the H2O AutoML to achieve better results. The Stacked Ensemble model is the best ML model using 100 and 200 features as shown in both tables. It is important to note that the Stacked Ensemble model that was built using the H2O AutoML with 200 features achieved the best performance in terms of Log Loss, true positive, and true negative metrics. It is a good combination because users want to avoid blocking their legitimate messages and ensure stopping spam messages.

Table \ref{tab1} - Table \ref{tab3} reveal that the ensemble methods achieved the best performance in all cases with the mljar-supervised AutoML and in two cases with the H2O AutoML tools. However, the best performance achieved using the TPOT AutoML is with the Logistic Regression model.

\begin{table}
\centering
\caption{The classification performance comparison using 50 features.}\label{tab1}
\begin{tabular}{|l|l|l|l|l|l|l|l|}
\hline
Tool &  Best model & Log Loss &  TP & FP & FN & TN & AUC\\ \hline
mljar-supervised & Stacked Ensemble & 1.2555 & 1085 & 42 & 9 & 267 & 0.9279\\\hline
H2O & Deep Learning& 7.6809 & \textbf{1091} & 309 & \textbf{3} & 0 & 0.4986 \\\hline
TPOT & Gradient Boosting& \textbf{1.1817} & 1087 & \textbf{41} & 7 & \textbf{268} & \textbf{0.9305}\\\hline
\end{tabular}
\end{table}

\begin{table}
\centering
\caption{The classification performance comparison using 100 features.}\label{tab2}
\begin{tabular}{|l|l|l|l|l|l|l|l|}
\hline
Tool &  Best model & Log Loss &  TP & FP & FN & TN & AUC \\ \hline
mljar-supervised & Stacked Ensemble & \textbf{0.9109} & \textbf{1087} & \textbf{30} & \textbf{7} & \textbf{279} & \textbf{0.9483} \\\hline
H2O & Stacked Ensemble& 1.0093 & 1084 & 31 & 10 & 278 & 0.9453 \\\hline
TPOT & Logistic Regression& 1.1817 & 1085 & 41 & 9 & 268 & 0.9295\\\hline
\end{tabular}
\end{table}

\begin{table}
\centering
\caption{The classification performance comparison using 200 features.}\label{tab3}
\begin{tabular}{|l|l|l|l|l|l|l|l|}
\hline
Tool &  Best model & Log Loss &  TP & FP & FN & TN & AUC\\ \hline
mljar-supervised & Stacked Ensemble & 0.8863 & \textbf{1088} & 30 & \textbf{6} & 279 & 0.9487 \\\hline
H2O & Stacked Ensemble& \textbf{0.8370} & \textbf{1088} & \textbf{28} & \textbf{6}  & \textbf{281} & \textbf{0.952} \\\hline
TPOT & Logistic Regression& 1.034 & 1086 & 34 & 8 & 275 & 0.9413\\\hline
\end{tabular}
\end{table}

With regards to the training time, as shown in Table \ref{time}, both the TPOT AutoML and mljar-supervised AutoML used the entire given time, while the H2O AutoML finished the training before the time-limit.

\begin{table}
\centering
\caption{Training time comparison. Time reported in H:MM format.}\label{time}
\begin{tabular}{|l|l|l|l|}
\hline
\multirow{2}{*}{Tool} & \multicolumn{3}{|c|}{Feature size} \\ \cline{2-4}
&  50 & 100 & 200\\\hline
mljar-supervised & 2:01 & 3:01 & 4:04\\\hline
H2O & 1:20 & 2:03 & 2:41 \\\hline
TPOT & 2:01 & 3:00 & 4:00\\\hline
\end{tabular}
\end{table}

\subsection{Best classification performance comparison}

As shown in Table \ref{tab4}, Stacked Ensemble models achieved the best performance with 100 and 200 features. As mentioned above and shown in Table \ref{tab5}, the Stacked Ensemble model that was built using the H2O AutoML with 200 features achieved the best classification performance. There is a 19.05\% improvement in Log Loss with respect to the TPOT AutoML and 5.56\% improvement with respect to the mljar-supervised AutoML. The detail of the obtained training results for the best model is shown in Table \ref{best}. It can also be seen in Table \ref{tab5} that the best performance was achieved with 200 features with all tools.

\begin{table}
\centering
\caption{The best classification performance comparison for each feature subset size.}\label{tab4}
\begin{tabular}{|l|l|l|l|}
\hline
Features subset size & Tool &  Best model & Log Loss \\ \hline
50 & TPOT & Gradient Boosting & 1.1817 \\\hline
100 & H2O & Stacked Ensemble& 1.0093 \\\hline
200 & H2O & Stacked Ensemble& \textbf{0.8370} \\\hline
\end{tabular}
\end{table}

\begin{table}
\centering
\caption{Best classification performance comparison for each tool.}\label{tab5}
\begin{tabular}{|l|l|l|l|l|}
\hline
Feature subset size & Tool &  Best model & Log Loss & Improvement (\%)\\ \hline
200 & TPOT & Logistic Regression& 1.034 &19.05\\\hline
200 & mljar-supervised & Ensemble & 0.8863 & 5.56\\\hline
200 & H2O & Stacked Ensemble& \textbf{0.8370} & - \\\hline
\end{tabular}
\end{table}

\begin{table}
\centering
\caption{Training results for the best model.}\label{best}
\begin{tabular}{|l|l|}
\hline
Metric &  Value\\ \hline
Log loss & 0.0766522\\\hline
TP & 3257\\\hline
FP & 24\\\hline
FN & 67\\\hline
TN & 859\\\hline
AUC & 0.994897\\\hline
Training time (in millisecond) & 2738 \\\hline
Prediction time per row (in millisecond) & 0.288793\\\hline
\end{tabular}
\end{table}

\section{Conclusions and future works}
In this work, the classification performance for SMS messages using three automatic ML tools was conducted. These tools are mljar-supervised AutoML, H2O AutoML, and TPOT AutoML. Three feature subset sizes were used with these tools. The main results of this work are summarized as follows:

\begin{itemize}
    \item The Stacked Ensemble model that was built using the H2O AutoML with 200 features achieved the best performance in terms of Log Loss, true positive, and true negative metrics. There is a 19.05\% improvement in Log Loss with respect to the TPOT AutoML tool and 5.56\% improvement with respect to the mljar-supervised AutoML tool. 
    \item Ensemble models (i.e., Stacked Ensemble and Gradient Boosting) achieved the best performance for each feature size.
    \item The best performance achieved with all tools was with 200 features.
\end{itemize}

The satisfactory filtering performance achieved with AutoML tools provides a potential application for AutoML tools to automatically determine the best ML model that can perform best for SMS spam message filtering. For future work, this work can be further extended by including more automatic ML tools, adding more features, and increasing training time-limit.

\section*{Acknowledgement}
The source code for this work is available in \url{https://github.com/Waddah-Saeed/SMS-Spam-Filtering-AutoML}.

\bibliographystyle{splncs04}
\bibliography{references}

\end{document}